\documentclass[sigconf, nonacm]{acmart}
\usepackage{graphicx} 
\usepackage{amsmath}
\usepackage{dsfont}
\usepackage{makecell}
\usepackage{booktabs}
\usepackage{subcaption}
\usepackage{siunitx}
\usepackage[most]{tcolorbox}
\usepackage{enumitem}

\usepackage{hyperref}

\newcommand{\huggingface}{\raisebox{-0.35ex}{\includegraphics[height=1.1em]{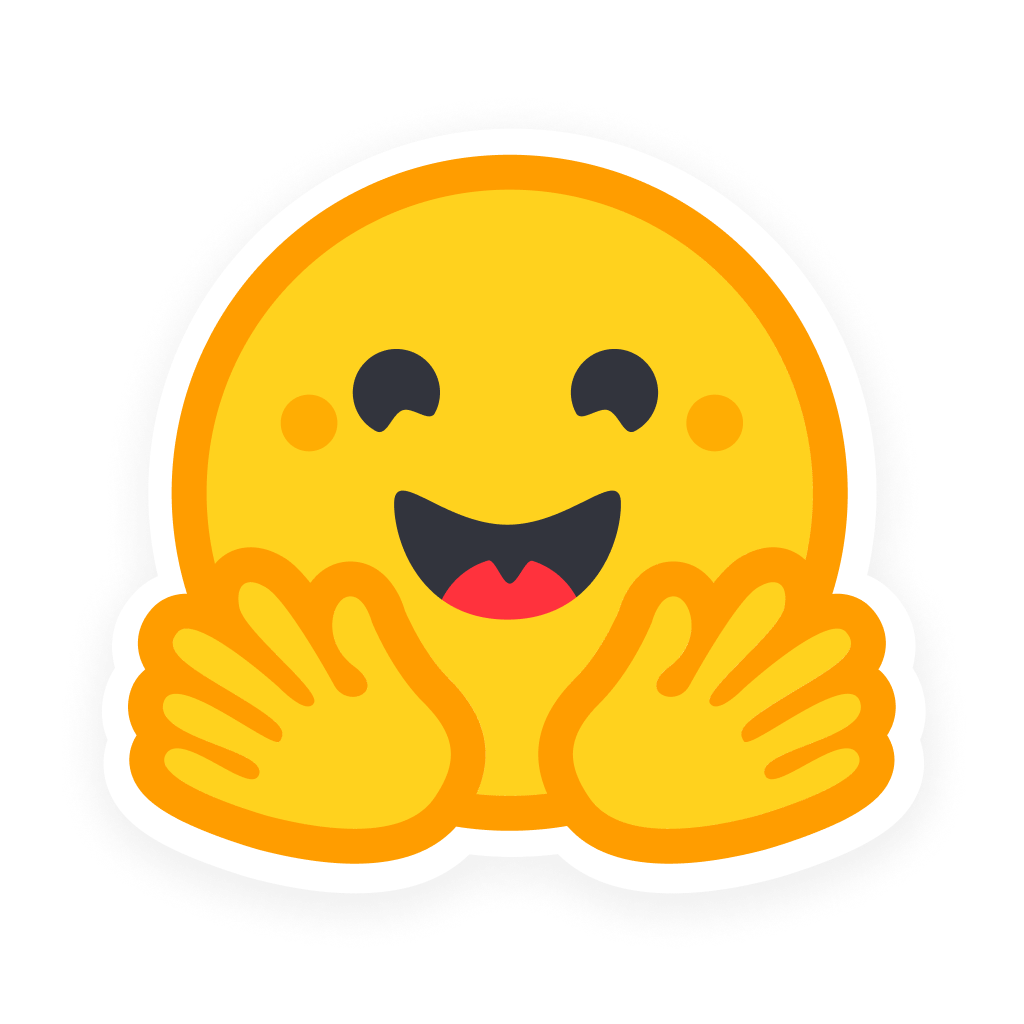}}}
\AtBeginDocument{%
  }

\setcopyright{none}                                  
\acmBooktitle{}
\acmYear{}
\acmISBN{}
\acmDOI{}
\renewcommand\footnotetextcopyrightpermission[1]{}   

\begin{document}

%
\title{SurgAtlas: A Large-Scale Surgical Video--Language Dataset with 2{,}391 Hours of Open and Minimally Invasive Surgery}


\author{Filippos Bellos}
\email{fbellos@umich.edu}
\affiliation{%
  \institution{University of Michigan}
  \city{Ann Arbor}
  \state{Michigan}
  \country{USA}
}

\author{Andre S. Gala-Garza}
\email{asgala@umich.edu}
\affiliation{%
  \institution{University of Michigan}
  \city{Ann Arbor}
  \state{Michigan}
  \country{USA}
}

\author{Miaowei Wang}
\email{m.wang-123@sms.ed.ac.uk}
\affiliation{%
  \institution{University of Edinburgh}
  \city{Edinburgh}
  \country{United Kingdom}
}

\author{Alyssa M. Hardin}
\email{alyssa.hardin@vumc.org}
\affiliation{%
  \institution{Vanderbilt University Medical Center}
  \city{Nashville}
  \state{Tennessee}
  \country{USA}
}

\author{Ahmad M. Hider}
\email{ahmad.hider@cuanschutz.edu}
\affiliation{%
  \institution{University of Colorado Anschutz Medical Campus}
  \city{Aurora}
  \state{Colorado}
  \country{USA}
}

\author{Li Yayuan}
\email{yayuanli@umich.edu}
\affiliation{%
  \institution{University of Michigan}
  \city{Ann Arbor}
  \state{Michigan}
  \country{USA}
}

\author{Jing Bi}
\email{jing.bi@rochester.edu}
\affiliation{%
  \institution{University of Rochester}
  \city{Rochester}
  \state{New York}
  \country{USA}
}

\author{Susan Liang}
\email{sliang22@rochester.edu}
\affiliation{%
  \institution{University of Rochester}
  \city{Rochester}
  \state{New York}
  \country{USA}
}

\author{Chenliang Xu}
\email{chenliang.xu@rochester.edu}
\affiliation{%
  \institution{University of Rochester}
  \city{Rochester}
  \state{New York}
  \country{USA}
}

\author{Donald S. Likosky}
\email{likosky@med.umich.edu}
\affiliation{%
  \institution{Michigan Medicine}
  \city{Ann Arbor}
  \state{Michigan}
  \country{USA}
}

\author{Jason J. Corso}
\email{jjcorso@umich.edu}
\affiliation{%
  \institution{University of Michigan}
  \city{Ann Arbor}
  \state{Michigan}
  \country{USA}
}

\renewcommand{\shortauthors}{Bellos et al.}


\begin{abstract}
We introduce SurgAtlas, the largest surgical video--language dataset to date, comprising 15{,}291 videos (2{,}391 hours) spanning 18 surgical specialties and over $5{,}000$ procedure types, sourced entirely from publicly available YouTube content. SurgAtlas is also the first surgical video--language dataset to include open surgery at scale, with 6{,}182 open procedure videos alongside over 9{,}000 minimally invasive recordings, and the first to establish standardized benchmarks for open-surgery video understanding. We additionally provide an expert-validated subset with verified visual question--answer pairs across diverse open and minimally invasive procedures, serving as a clinically grounded benchmark for surgical reasoning. Compared with existing surgical video--language datasets, SurgAtlas provides one of the most diverse annotation schemas, combining segment-level captions, step- and phase-level descriptions, video-level surgical descriptions, and reasoning-oriented question--answer pairs organized within a hierarchical taxonomy. These annotations are constructed through an automated multi-tier pipeline with LLM-based enrichment and a staged VQA generation framework with explicit groundedness verification. The scale and diversity of SurgAtlas enable training surgical foundation models with broad procedural coverage: we finetune Qwen3-VL-8B through a two-stage captioning-then-instruction pipeline and achieve competitive or state-of-the-art results on multiple established surgical benchmarks, including phase recognition, triplet detection, and reasoning question answering. More broadly, SurgAtlas provides a large native public video corpus that can support future large-scale pretraining of multimodal surgical AI systems and contribute to the development of next-generation foundation models for surgery. Dataset publicly available at
\href{https://huggingface.co/datasets/filbel/SurgAtlas}{\huggingface\texttt{filbel/SurgAtlas}}.
\end{abstract}



\keywords{
video-language dataset, surgical video understanding, open surgery, multimodal benchmark, vision-language models, visual question answering, medical AI
}

\begin{teaserfigure}
    \includegraphics[width=\textwidth]{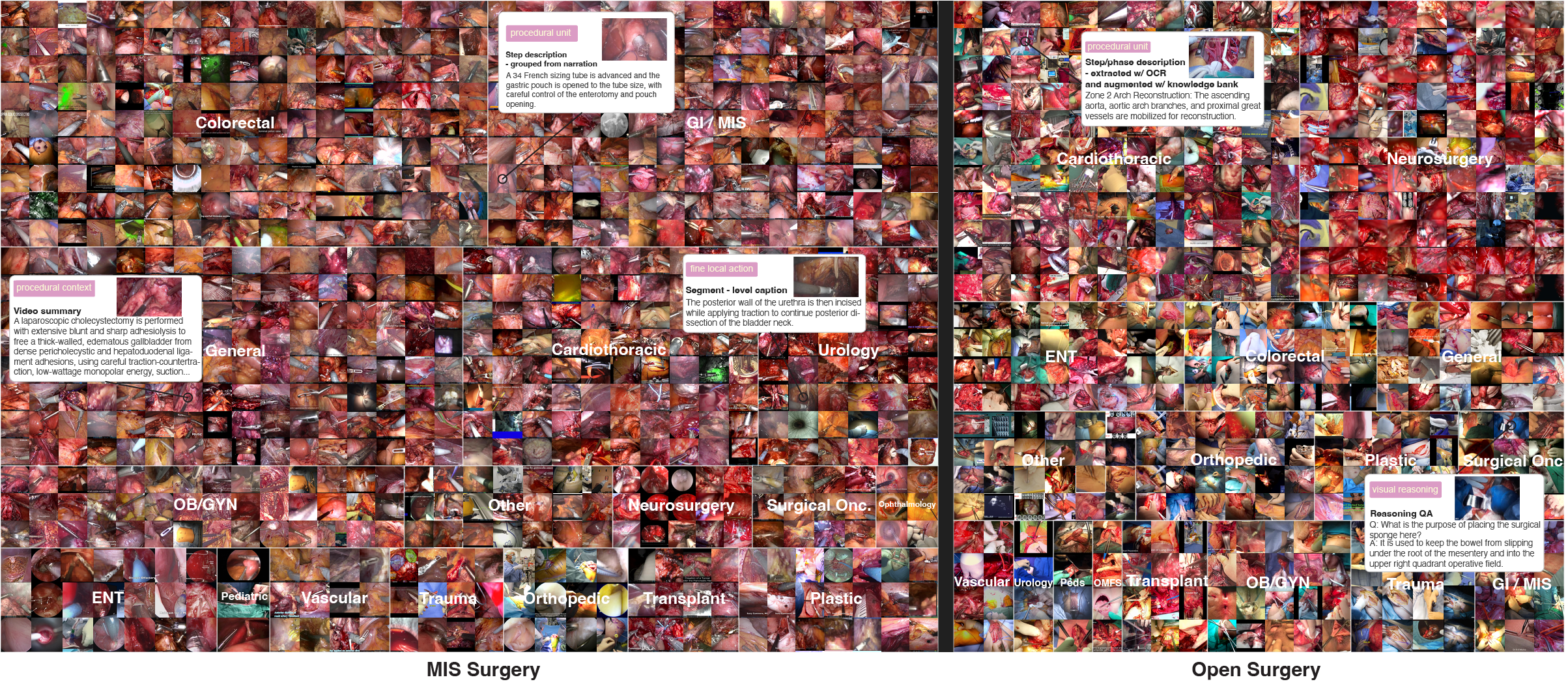}
\caption{\textbf{Visual overview of SurgAtlas.} Sampled frames organized by surgical specialty and split into minimally invasive (left) and open surgery (right). Insets illustrate the diverse annotation types that SurgAtlas uniquely combines in a single corpus.}
    \label{fig:statement_1}
\end{teaserfigure}

\maketitle


\section{Introduction}

Large multimodal foundation models have shown strong performance across domains by aligning visual content with natural language, sparking growing interest in surgical vision--language tasks where understanding must extend beyond static recognition to procedural context, temporal structure, and clinically meaningful reasoning. Early efforts learned alignment from narrated academic videos through contrastive supervision~\cite{SurgVLP,HecVL,PeskaVLP,SurgLaVi}, and more recent generative systems have moved toward open-ended question answering and conversational reasoning~\cite{SUREON,SurgVLM,SurgSigma,SurgLLM,EndoChat}. Despite these advances, three limitations continue to constrain the field.

\textit{First, supervision sources remain narrow.} Recent large-scale generative systems achieve scale primarily by reformatting structured annotations from public datasets into multimodal conversations~\cite{SurgVLM,SurgSigma}; while effective for sample count, this strategy is bounded by the closed-vocabulary ontologies of the source datasets and cannot capture why a maneuver is performed, what anatomical risk it mitigates, or what step should follow. SUREON~\cite{SUREON} moves beyond ontology conversion by deriving QA pairs from expert narration, but narration is pedagogically selective: narrators emphasize teaching points and decisions while routine operative steps receive comparatively little commentary. \textit{Second, the underlying public surgical video corpus is small.} Existing narration-centered resources rely on a few thousand source videos---approximately 1.3K in SurgVLP~\cite{SurgVLP} (largely from access-restricted platforms) and 2{,}464 in the public portion of SurgLaVi~\cite{SurgLaVi}, which also underlies SUREON. Recent integrated systems do not resolve this: Surg$\Sigma$-DB and SurgVLM-DB derive from a pool of only 1.59K unique sources, partly in-house and partly from existing ontology-based public datasets~\cite{SurgSigma,SurgVLM}. 
\textit{Third, open surgery is largely absent.} Existing public surgical VL datasets are overwhelmingly minimally invasive, endoscopic, or otherwise scoped~\cite{SurgVLP,PeskaVLP,SurgLaVi,SUREON,SurgVLM,SurgSigma}. The two prior open-surgery resources fall an order of magnitude short of what language supervision requires: AVOS~\cite{AVOS} provides only 47 hours from 343 of its 1{,}997 videos and limits annotations to bounding boxes and action labels, and EgoSurgery~\cite{EgoSurgery} contains 20 egocentric videos with 9 categorical phases. Neither offers any language supervision. 
Yet open surgery accounts for a major share of operative care (e.g., cardiothoracic, transplant, orthopedic) and presents a fundamentally different visual regime: wide operative fields, multiple operators' hands, directly exposed anatomy, and ambient lighting rather than an endoscopic field with a small number of instruments. To our knowledge, no prior work has established a large-scale public surgical video--language dataset including open surgery at scale together with a standardized benchmark for open-surgery understanding.

We address these limitations with \textbf{SurgAtlas}, a surgical video--language dataset built entirely from publicly available YouTube content, comprising 15{,}291 videos and 2{,}391 hours across 18 specialties and over 5{,}000 procedures, including 6{,}182 open-surgery videos alongside over 9{,}000 minimally invasive recordings. Unlike systems whose breadth comes from integrating existing public datasets or converting structured annotations~\cite{SurgVLM,SurgSigma}, SurgAtlas derives its core scale from direct collection of raw surgical videos paired with multi-source supervision extraction. The pipeline recovers different types of supervision from different signals: refined sentence-level captions from narration and step descriptions that group narrated segments into procedure- and phase-level units rather than just didactic commentary (Tier~1), 
phase and step labels expanded into procedural descriptions for videos with informative on-screen text (Tier~2), and procedure-level overviews from metadata-only videos (Tier~3). On top of these grounded annotations, we generate reasoning-oriented question--answer pairs organized in a hierarchical taxonomy. We further establish the first standardized benchmark for open-surgery video understanding, including an expert-validated subset that provides a clinically reviewed resource for trustworthy evaluation.
To validate the resulting corpus, we finetune Qwen3-VL-8B with a two-stage captioning-then-instruction recipe and demonstrate competitive or state-of-the-art performance across various surgical video understanding tasks.

\section{Related Work}

\begin{figure*}[t]
    \centering
    \includegraphics[width=\textwidth]{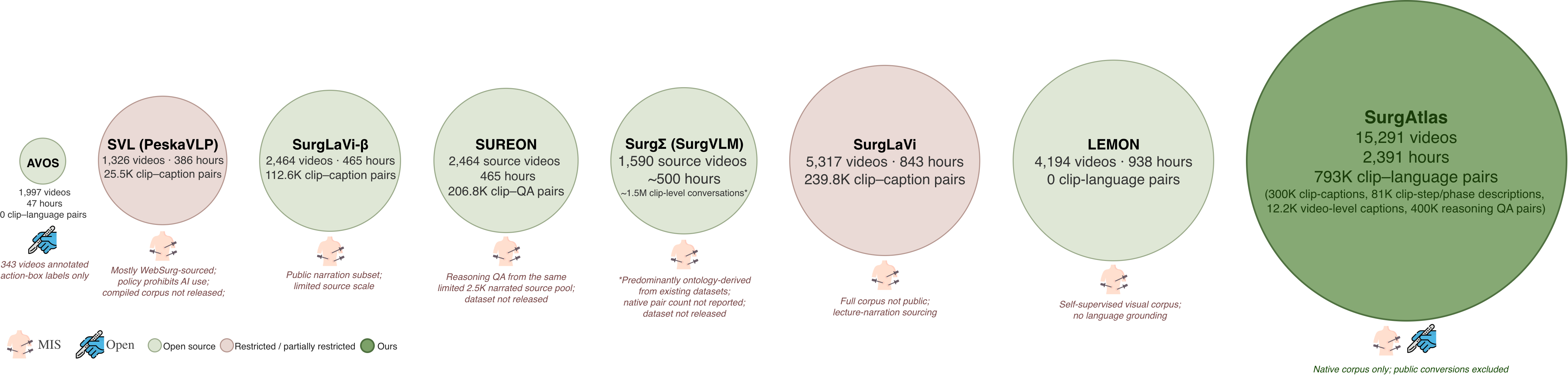}
\caption{\textbf{Comparison of large-scale surgical video datasets.}
Bubble area is proportional to video duration; color denotes
public availability, and icons indicate minimally invasive and open-surgery
coverage. Bubble text reports source videos, duration, and video/clip-level
language annotations. For Surg$\Sigma$, $\sim$1.5M denotes clip-level
conversations; its 5.98M total also includes image-level data. Callouts
summarize limitations in supervision, source scale, or release status.
SurgAtlas spans both surgical regimes
and provides 793K multi-granular video/clip--language annotations, excluding
the separately converted public datasets.}
    \label{fig:scale}
\end{figure*}

\paragraph{Surgical video datasets.}
Early surgical video datasets provided dense frame-level annotations for individual procedures (Cholec80~\cite{Cholec80}, CholecT50~\cite{CholecT50}, AutoLaparo~\cite{AutoLaparo}, MultiBypass140~\cite{MultiBypass140}, SAR-RARP50~\cite{SAR-RARP50}, Cataract-101~\cite{Cataract-101}). These benchmarks remain invaluable for phase, action, and workflow analysis but are small, narrow, and limited to categorical labels. More recent efforts (LEMON~\cite{LEMON}, SurgBench~\cite{SurgBench}, SurgVISTA~\cite{SurgVISTA}, GenSurgery~\cite{GenSurgery}) substantially expand visual scale for self-supervised pretraining but provide no language grounding. All of the above are restricted to minimally invasive surgery.

\paragraph{Surgical vision--language datasets and models.}
Two main directions have emerged. \emph{Narration-centered} datasets pair surgical videos with expert spoken commentary: SurgVLP~\cite{SurgVLP} introduced ASR-based dual-encoder pretraining; HecVL~\cite{HecVL} and PeskaVLP~\cite{PeskaVLP} added hierarchical and LLM-augmented supervision; SurgLaVi~\cite{SurgLaVi} scaled to a hierarchical clip--caption corpus; and SUREON~\cite{SUREON} demonstrated that reasoning-oriented VQA can be harvested from the same narrated videos. \emph{Integrated multimodal corpora} take a different route: SurgVLM~\cite{SurgVLM} and Surg$\Sigma$~\cite{SurgSigma} consolidate diverse public datasets into unified instruction-tuning collections, and SurgLLM~\cite{SurgLLM} and EndoChat~\cite{EndoChat} target spatial and conversational reasoning. The first direction is bounded by the scale of available narrated lectures and by narration's pedagogical selectivity; the second derives scale primarily from converting ontology-based labels into conversations rather than from native video--language supervision. Neither line treats open surgery as a first-class setting.

\section{SurgAtlas Dataset}
\subsection{Overview}

Each video in SurgAtlas carries multigranular textual annotations across four levels of abstraction (Figure~\ref{fig:pipeline}). \textit{Segment-level captions} (300K pairs) provide fine-grained descriptions of individual surgical actions, each paired with a clip. \textit{Step- and phase-level descriptions} (81K pairs) capture coherent procedural units---56K formed by grouping adjacent narrated segments in Tier~1 and 25K from canonicalized on-screen phase or step labels in Tier~2. \textit{Video-level summaries} (12.2K) are concise 2--3 sentence clinical case overviews constrained to visually grounded content, derived from narration in Tier~1 (9.4K) and from metadata in Tier~3 (2.8K). \textit{Reasoning QA pairs} (400K) are hierarchically organized across five broad and ten fine-grained categories (Section~\ref{sec:taxonomy}) and produced through a staged pipeline with explicit groundedness verification. Rather than forcing all videos into a single supervision template, the pipeline extracts the richest available signal from each source, yielding annotations that span fine-grained local action grounding through higher-level procedural structure and reasoning supervision.

\subsection{Dataset Construction Pipeline}
\label{sec:pipeline}

\begin{figure*}[t]
    \centering
    \includegraphics[width=\textwidth]{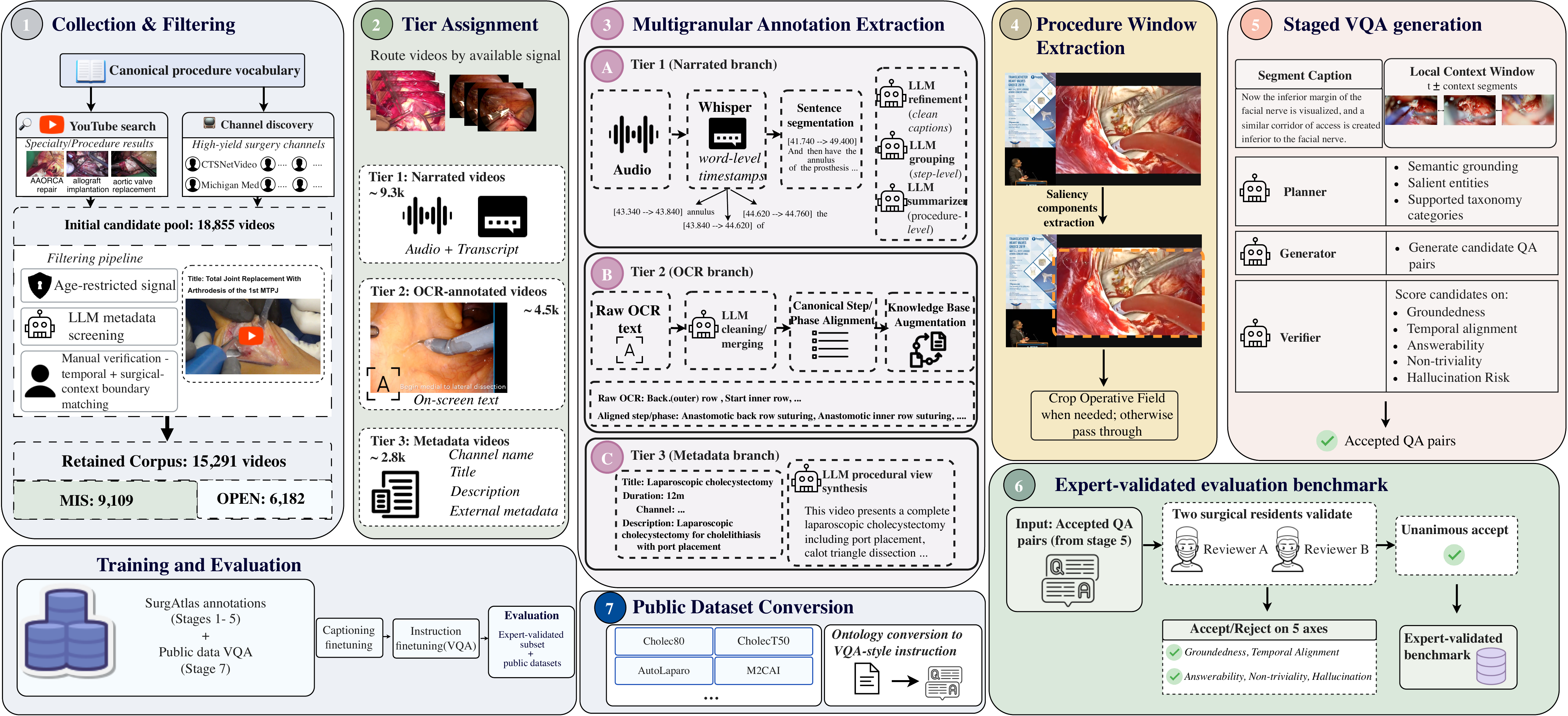}
\caption{\textbf{Overview of the SurgAtlas construction pipeline.} (1) YouTube search and channel discovery yield $18{,}855$ candidates,
of which $15{,}291$ are retained ($9{,}109$ MIS / $6{,}182$ open).
(2) Videos are routed to narration-, OCR-, or metadata-based annotation
tiers. (3) Tier-specific extraction produces segment-, step-, and
procedure-level annotations through Whisper and LLM refinement,
OCR cleaning and canonical-step alignment, or metadata-driven synthesis.
(4) Presentation videos are cropped to the operative field.
(5) A Planner--Generator--Verifier loop generates VQA pairs scored on
5 quality axes. (6) Surgical residents validate a subset to form
the expert benchmark $\mathcal{A}^{\mathrm{exp}}$.
(7) Public datasets are converted to VQA-style supervision through
ontology mapping.}
    \label{fig:pipeline}
\end{figure*}

We denote the SurgAtlas corpus by $\mathcal{D} = \{(v_i, \mathcal{A}_i)\}_{i=1}^{N}$, where $v_i$ is the $i$-th surgical video and $\mathcal{A}_i$ is its associated multigranular annotation set. Each video $v_i$ is a sequence of frames $v_i = (f_i^{1}, \ldots, f_i^{T_i})$ with duration $\tau(v_i)$, and is associated with three optional auxiliary signals: an audio track $a_i$, a sequence of on-screen text overlays $\mathcal{O}_i$, and metadata $m_i = (\text{title}_i, \text{description}_i, \text{channel}_i)$. The pipeline cascades over these signals, extracting the richest available supervision per video (Fig.~\ref{fig:pipeline}).
\subsubsection{Stage 1: Collection and Filtering}

We define a canonical surgical procedure vocabulary $\mathcal{P} = \{p_1, \ldots, p_{|\mathcal{P}|}\}$ obtained from the American College of Surgeons specialty taxonomy and the AMA CPT codes~\cite{CPT}, spanning $18$ surgical specialties. Collection proceeds along two complementary axes. First, search queries are issued for each $p \in \mathcal{P}$ and progressively expanded into specialty-specific variants. Returned candidates are matched against $\mathcal{P}$ via a token-level similarity score $\sigma(\text{title}_j, p) = |\mathcal{T}(\text{title}_j) \cap \mathcal{T}(p)| / |\mathcal{T}(\text{title}_j) \cup \mathcal{T}(p)|$, where $\mathcal{T}(\cdot)$ denotes the set of normalized lemmatized tokens, and we retain candidates with $\max_{p \in \mathcal{P}} \sigma(\text{title}_j, p) \geq \theta_{\sigma}$. This Jaccard-style matching captures lexical variations of the same procedure (e.g., ``laparoscopic cholecystectomy'' vs.\ ``cholecystectomy'') beyond exact keyword matches. Second, we identify high-yield channels---university surgery departments, professional societies, and surgical education platforms---as systematic sources, since such channels typically host complete operative libraries that span multiple procedure types within a single specialty.

The combined yield forms a candidate pool $\mathcal{C}_{\text{init}} = \{v_j\}_{j=1}^{N_{\text{init}}}$ with $N_{\text{init}} = 18{,}855$.\,
We then apply a sequence of filtering operators $\Phi = \Phi_{\text{man}} \circ \Phi_{\text{LLM}} \circ \Phi_{\text{age}}$, where $\Phi_{\text{age}}$ retains age-restricted videos as a positive signal: YouTube's content policies systematically flag real intraoperative footage as age-restricted, making restriction status a reliable indicator of authentic operative content. Next, $\Phi_{\text{LLM}}$ is a screening operator $\phi_{\text{meta}}: m_j \mapsto \{0,1\}$ that jointly inspects the title and description $(\text{title}_j, \text{description}_j)$ to classify whether the video contains operative footage in any format---raw intraoperative recordings, case demonstrations, or conference-style presentations of surgical cases---and filters out content that is purely didactic without operative footage (lectures, panel discussions, grand rounds). Finally, $\Phi_{\text{man}}$ denotes human verification of operative content by 17 annotators, together with annotation of the temporal boundaries $\mathcal{B}_j = \{(s_j^{(k)}, e_j^{(k)})\}_k$ delimiting the surgical portion of each video and a coarse surgery-type label (open vs. minimally invasive); in the same pass, annotators flag videos in presentation or conference format for downstream procedure window extraction (Stage 4), distinguishing them from raw full-frame intraoperative recordings that require no spatial cropping.
The retained corpus is $\mathcal{C} = \Phi(\mathcal{C}_{\text{init}})$ with $|\mathcal{C}| = N = 15{,}291$, of which $N_{\text{open}} = 6{,}182$ are open-surgery procedures and $N_{\text{MIS}} =9{,}109$ are minimally invasive.
\subsubsection{Stage 2: Tier Assignment}
Each video $v_i \in \mathcal{C}$ is routed to one or more annotation pipelines based on its available linguistic signals: \textbf{Tier~1 (narrated)}, with expert spoken commentary; \textbf{Tier~2 (OCR-annotated)}, with on-screen procedural text indicating phases, steps, or instruments; and \textbf{Tier~3 (metadata-only)}, the residual case where only title, description, and source channel are available. Tier assignment uses two binary indicators $\mathbb{1}_{\mathrm{narr}}(v_i)$ and $\mathbb{1}_{\mathrm{ocr}}(v_i)$, initialized by a speech-activity classifier and OCR temporal coverage respectively, then verified by manual annotators. The tier-membership function $\rho: \mathcal{C} \to 2^{\{1,2,3\}}$ assigns $v_i$ to Tier~1 if $\mathbb{1}_{\mathrm{narr}}(v_i)=1$, to Tier~2 if $\mathbb{1}_{\mathrm{ocr}}(v_i)=1$, and to Tier~3 otherwise. Tiers~1 and~2 may overlap; Tier~3 is the strict residual $\mathcal{C} \setminus (\mathcal{C}_1 \cup \mathcal{C}_2)$. The resulting sizes are $|\mathcal{C}_1| \approx 9{,}360$, $|\mathcal{C}_2| \approx 4{,}460$, $|\mathcal{C}_3| \approx 2{,}778$, with $|\mathcal{C}_1 \cap \mathcal{C}_2| \approx 1{,}307$.

\subsubsection{Stage 3: Multigranular Annotation Extraction}

For every $v_i$ we produce an annotation set $\mathcal{A}_i = \{\mathcal{A}_i^{\text{seg}}, \mathcal{A}_i^{\text{step}}, \mathcal{A}_i^{\text{vid}}, \mathcal{A}_i^{\text{qa}}\}$ corresponding to the four levels of granularity. The exact extractors are tier-dependent.

\textbf{Tier 1 (narration).} Whisper~\cite{Whisper} is applied to the audio track to produce a word-level transcript $\mathcal{W}_i = \{(w_i^{(j)}, t_i^{(j),s}, t_i^{(j),e})\}_{j=1}^{J_i}$ with per-word start and end timestamps. We deliberately discard Whisper's native segmentation: its segments routinely cut a single explanatory utterance into several short fragments, breaking the surgeon's reasoning across boundaries. We instead reconstruct sentence-level units by aggregating consecutive words into coherent sentences, with each sentence inheriting its start time from its first word and its end time from its last word, yielding $\mathcal{S}_i = \{(t_i^{(k),s}, t_i^{(k),e}, x_i^{(k)})\}_{k=1}^{K_i}$ in which each $x_i^{(k)}$ corresponds to a complete narrated thought rather than an arbitrary acoustic chunk.

From this sentence sequence we construct annotations at three increasing levels of abstraction, from local action descriptions to whole-procedure overviews. \emph{(i) Segment-level captions} (finest). A language-model refinement operator $\mathcal{R}_{\text{LLM}}: x_i^{(k)} \mapsto c_i^{(k)}$ rewrites each sentence into a clean, terminology-corrected description of the surgical activity narrated over the corresponding interval $[t_i^{(k),s}, t_i^{(k),e}]$, and a relevance gate $g_{\text{rel}}(c_i^{(k)}) \in \{0,1\}$ retains only sentences that describe operative content---excluding speaker introductions, informal speech, discussion not pertaining to the operative field, and closing remarks---yielding $\mathcal{A}_i^{\text{seg}} = \{(v_i[t_i^{(k),s}, t_i^{(k),e}], c_i^{(k)}) : g_{\text{rel}}(c_i^{(k)}) = 1\}$. \emph{(ii) Step-level descriptions} (intermediate). An LLM grouping operator $\Psi_{\text{LLM}}$ takes the timestamped, refined sentences $\{(t_i^{(k),s}, t_i^{(k),e}, c_i^{(k)})\}_k$ together with the video metadata $m_i$ and partitions them into procedural steps by merging consecutive sentences that share a common surgical goal: $\mathcal{A}_i^{\text{step}} = \{(v_i[t_i^{(\alpha),s}, t_i^{(\beta),e}], \mathcal{R}_{\text{LLM}}^{\text{step}}(c_i^{(\alpha:\beta)})) : (\alpha, \beta) \in \Psi_{\text{LLM}}(\mathcal{S}_i, m_i)\}$. Conditioning on $m_i$ allows the grouping to align with the procedure-specific phase ontology rather than relying on lexical cues alone. \emph{(iii) Video-level summaries} (coarsest). A constrained operator $\mathcal{R}_{\text{LLM}}^{\text{vid}}$ produces a whole-procedure overview from the full sentence sequence and metadata, $\mathcal{A}_i^{\text{vid}} = \mathcal{R}_{\text{LLM}}^{\text{vid}}(c_i^{(1:K_i)}, m_i)$, with an explicit prompt constraint excluding patient history, preoperative imaging, and postoperative plans so that the summary stays grounded in visible operative content.

\textbf{Tier 2 (on-screen procedural text).} OCR yields a raw label sequence $\mathcal{O}_i^{\text{raw}} = \{(\ell_i^{(k)}, t_i^{(k),s}, t_i^{(k),e})\}_{k=1}^{L_i}$ that contains substantial non-procedural noise: instrument-vendor logos, surgeon and institution attributions, and frame-to-frame duplicates of the same overlay. We apply an LLM-based cleaning operator $\mathcal{R}_{\text{LLM}}^{\text{ocr}}$, conditioned on the procedure type $p_i$, that for each entry either returns a normalized procedural label or marks it as non-procedural; surviving entries that share the same normalized label across temporally adjacent intervals are then merged via a temporal consolidation operator $\Gamma$, which additionally extends the end time of each kept segment to the start of the next surgically relevant segment when the intervening gap contains operative content. 
The resulting cleaned sequence is $\mathcal{O}_i = \Gamma\big(\mathcal{R}_{\text{LLM}}^{\text{ocr}}(\mathcal{O}_i^{\text{raw}} \mid p_i)\big) = \{(\tilde{\ell}_i^{(k)}, t_i^{(k),s}, t_i^{(k),e})\}_k$. For each procedure we generate a procedure-conditioned canonical step list (a knowledge base) $\mathcal{K}(p) = (k_1^{p}, \ldots, k_{M_p}^{p})$ via $\mathcal{R}_{\text{LLM}}^{\text{kb}}: p \mapsto \mathcal{K}(p)$, where each entry pairs a canonical step name with a procedure-specific description. Each cleaned label is aligned to this list via $\hat{\ell}_i^{(k)} = \arg\max_{k' \in \mathcal{K}(p_i)} \text{sim}(\mathbf{e}(\tilde{\ell}_i^{(k)}), \mathbf{e}(k'))$ subject to $\text{sim} \geq \theta_s$, where $\mathbf{e}(\cdot)$ is a sentence embedding. When the match satisfies the threshold, the matched step is expanded into its procedure-specific descriptive passage $d_i^{(k)} = \mathcal{R}_{\text{LLM}}^{\text{exp}}(\hat{\ell}_i^{(k)} \mid p_i)$; otherwise we retain the normalized OCR label $\tilde{\ell}_i^{(k)}$ unchanged---a conservative \emph{phase fallback} that preserves the clinically authored on-screen signal and limits the introduction of unsupported procedural detail when canonical-step alignment is uncertain. This yields $\mathcal{A}_i^{\text{step}} = \{(v_i[t_i^{(k),s}, t_i^{(k),e}], \tau_i^{(k)})\}_k$, where $\tau_i^{(k)} = d_i^{(k)}$ if the label was matched and $\tau_i^{(k)} = \tilde{\ell}_i^{(k)}$ otherwise (see Appendix~\ref{app:construction_examples} for the fallback rate and worked examples).

\textbf{Tier 3 (metadata-only).} For videos with neither narration nor structured OCR, we set $\mathcal{A}_i^{\text{vid}} = \mathcal{R}_{\text{LLM}}^{\text{meta}}(m_i)$, where $\mathcal{R}_{\text{LLM}}^{\text{meta}}$ generates a procedural overview from the title, description, and source channel, and leave finer-granularity annotations empty.

\subsubsection{Stage 4: Procedure Window Extraction}

For each $v_i$ flagged by $\Phi_{\text{man}}$ as presentation- or conference-format, we estimate a spatial crop $\mathbf{b}_i = (x_i, y_i, w_i, h_i)$ that isolates the operative content from presentation overlays such as speaker panels, institutional logos, and slide chrome. We sample frames $\{f_i^{(\tau)}\}_{\tau \in \mathcal{T}_s}$ uniformly across the surgical portion of the video and define a composite saliency map $S_i(x,y) = \mathbb{E}_{\tau \in \mathcal{T}_s}[\lambda_1\,S^{\text{sat}}_i(x,y;\tau) + \lambda_2\,S^{\text{edge}}_i(x,y;\tau) + \lambda_3\,S^{\text{flow}}_i(x,y;\tau)]$, combining color saturation, edge magnitude, and inter-frame optical flow---three cues that are jointly high inside the operative field (saturated tissue, fine instrument edges, manipulation-induced motion) and jointly low across static presentation chrome. The bounding box is recovered as $\mathbf{b}_i = \text{bbox}(\text{LCC}(\mathds{1}[S_i \geq \tau_S] \oplus \mathcal{M}))$, where $\oplus$ denotes morphological closing with kernel $\mathcal{M}$ and LCC extracts the largest connected component. The crop is applied iff $\min(w_i / W_i, h_i / H_i) \leq 1 - 0.04$, avoiding unnecessary re-encoding of videos whose operative content already fills the frame. Videos not flagged in Stage~1 are passed through unmodified.

\subsubsection{Stage 5: Staged VQA Generation}
\label{sec:taxonomy}

To produce reasoning-level supervision beyond descriptive captioning, we transform segment-level annotations into grounded visual question--answer pairs through a staged generation pipeline. Each pair is generated locally from a single segment-level caption together with its temporal neighborhood, ensuring that supervision remains anchored to a specific operative moment rather than the procedure as a whole.

QA pairs are organized in a two-level taxonomy $\mathcal{Y} = \{(y_b, y_f) : y_b \in \mathcal{Y}_B,\, y_f \in \mathcal{Y}_F(y_b)\}$ with $|\mathcal{Y}_B| = 5$ broad categories and $|\mathcal{Y}_F| = 10$ fine-grained subcategories. The broad categories are \textit{perception and identification}, \textit{action and procedural state}, \textit{operative reasoning}, \textit{temporal and predictive reasoning}, and \textit{risk anatomy identification}. Perception subsumes \textit{entity existence}, \textit{entity state}, and \textit{spatial relation}, with \textit{entity state} covering both anatomical states (exposure, dissection plane, tension) and observable instrument-state transitions (clip deployment, energy activation) that are difficult to hallucinate without grounding. Action and procedural state contains \textit{instrument--tissue interaction} and \textit{operative action}, the latter spanning generic visual action and procedure-specific maneuvers. Operative reasoning splits into \textit{maneuver rationale}, the immediate clinical justification for the current action, and \textit{decision justification}, higher-level operative choices such as approach selection or conversion. Temporal and predictive reasoning spans \textit{procedural sequence}, unifying summarization and ordering over a short temporal window, and \textit{next-step prediction}. \textit{Risk anatomy identification} covers anatomical structures requiring protection and explicit safety practices such as the critical view of safety. The taxonomy was developed with surgical collaborators to emphasize categories with strong visual grounding and reduced semantic overlap.
\paragraph{Pipeline.}
For each segment-level pair $(v_i[t_i^{(k),s}, t_i^{(k),e}], c_i^{(k)}) \in \mathcal{A}_i^{\text{seg}}$, we form the local context $\mathcal{C}_i^{(k)} = (c_i^{(k-w:k+w)}, m_i)$ comprising a window of $2w$ neighboring captions and the video metadata, and apply three LLM operators in sequence. The \emph{planner} produces $\pi_i^{(k)} = \mathcal{R}_{\text{plan}}(c_i^{(k)}, \mathcal{C}_i^{(k)})$, returning a tuple $\pi_i^{(k)} = (\sigma_i^{(k)}, \mathcal{E}_i^{(k)}, \mathcal{Y}_i^{(k)})$ in which $\sigma_i^{(k)}$ is a one-sentence summary of the semantic grounding moment, $\mathcal{E}_i^{(k)}$ is the set of salient entities visible in the clip, and $\mathcal{Y}_i^{(k)} \subseteq \mathcal{Y}$ is the set of taxonomy categories supported by the available evidence---the planner is constrained to select at most $K_{\text{cat}}$ fine-grained categories whose broad parents are jointly consistent. The \emph{generator} then produces candidate pairs $\mathcal{Q}_i^{(k)} = \mathcal{R}_{\text{gen}}(c_i^{(k)}, \mathcal{C}_i^{(k)}, \pi_i^{(k)}) = \{(q, a, y_f, \texttt{evidence})\}$ with one candidate per selected fine-grained category, where each candidate carries a pointer to the supporting evidence in $c_i^{(k)} \cup \mathcal{C}_i^{(k)}$. The \emph{verifier} yields $\tilde{\mathcal{Q}}_i^{(k)} = \mathcal{R}_{\text{judge}}(\mathcal{Q}_i^{(k)}, c_i^{(k)}, \mathcal{C}_i^{(k)})$ by scoring each candidate along five axes---groundedness $\gamma_g$, temporal alignment $\gamma_t$, answerability $\gamma_a$, non-triviality $\gamma_n$, and (inverted) hallucination risk $\gamma_h$---each on an integer scale $\{1, \ldots, 5\}$. A candidate is accepted iff $\min(\gamma_g, \gamma_t, \gamma_a, \gamma_n, \gamma_h) \geq \theta_q$ and may be lightly revised in question or answer form by the verifier without altering meaning. The reasoning annotation set is the union over accepted candidates, $\mathcal{A}_i^{\text{qa}} = \bigcup_k \tilde{\mathcal{Q}}_i^{(k)}$.

\paragraph{Question format.}
Each accepted pair $(q, a, y_f) \in \mathcal{A}_i^{\text{qa}}$ is generated in either open-ended or multiple-choice form: open-ended pairs yield $a$ as a free-form answer of one to two sentences; multiple-choice pairs additionally produce four candidate options $(o_1, o_2, o_3, o_4)$ with a designated correct index, drawn so that distractors are clinically plausible but unambiguously incorrect from the clip evidence. This dual-format design supports both generative training and discriminative evaluation from the same annotation source. Representative examples spanning all ten fine-grained categories are provided in Figure~\ref{fig:vqa_examples}.

\subsubsection{Stage 6: Expert-Validated Evaluation Subset}
\label{sec:expert_subset}

To complement the automatically generated reasoning supervision in $\mathcal{A}_i^{\text{qa}}$ with a clinically validated resource, we construct a held-out subset $\mathcal{A}^{\text{exp}} \subset \bigcup_i \mathcal{A}_i^{\text{qa}}$ through independent expert review. Two surgical residents independently evaluated $N_{\text{rev}}=2,960$ candidate question--answer pairs sampled proportionally to surgical-specialty distribution and across the full taxonomy, covering both open and minimally invasive procedures, judging each on a binary accept/reject scale ($\{0, 1\}$) along the same five axes used by the Stage~5 verifier: groundedness, temporal alignment, answerability, non-triviality, and absence of unsupported clinical claims. A pair was retained in $\mathcal{A}^{\text{exp}}$ only if both reviewers independently accepted it on all five axes; disagreements were dropped. Because LLM-generated supervision is known to carry systematic biases and hallucinations, this two-reviewer unanimous-accept criterion gives a clinically validated subset that does not depend on any LLM in the loop. We use $\mathcal{A}^{\text{exp}}$ purely for evaluation---no pair or clip in $\mathcal{A}^{\text{exp}}$ appears in any training stage---and report it alongside the automatically generated test split.

\subsubsection{Stage 7: Public Dataset Conversion}

For each public source dataset $\mathcal{D}_j^{\text{pub}}$, structured annotations are mapped into VQA-style supervision via a dataset-specific template operator $\mathcal{T}_j: (f, y) \mapsto (q, a)$. We convert the training splits of 18 public surgical datasets covering phase, action, triplet, instrument, anatomy, and safety supervision: Cholec80~\cite{Cholec80}, 
CholecT50~\cite{CholecT50}, HeiChole~\cite{HeiChole}, AutoLaparo~\cite{AutoLaparo}, MultiBypass140~\cite{MultiBypass140}, M2CAI16~\cite{M2CAI16}, SAR-RARP50~\cite{SAR-RARP50}, Cataract1K~\cite{Cataract1K}, EndoVis-2017 and EndoVis-2018~\cite{EndoVis2017,EndoVis2018}, LapGyn4~\cite{LapGyn4}, GraSP~\cite{GraSP}, CholecSeg8k~\cite{CholecSeg8k}, Endoscapes-2023 (general and CVS subsets)~\cite{Endoscapes2023}, PitVQA~\cite{PitVQA}, PhaKIR~\cite{PhaKIR}, DSAD~\cite{DSAD}, and SurgVU~\cite{SurgVU}. Only training splits are used: $\mathcal{D}^{\text{IT}} = \bigcup_j \mathcal{T}_j(\mathcal{D}_j^{\text{pub,train}})$, with test splits held out for downstream evaluation.

\subsection{Dataset Analysis}
\label{sec:dataset_analysis}

We analyze SurgAtlas along four axes: scale, diversity, richness, and quality. 

\paragraph{Scale.}
SurgAtlas contains $N = 15{,}291$ surgical videos and $\tau(\mathcal{D}) = 2{,}391$ hours of operative footage, sourced entirely from publicly available YouTube content. To our knowledge, this constitutes the largest native public video corpus in surgical vision--language learning, exceeding the publicly released portion of SurgLaVi~\cite{SurgLaVi} ($\sim$2{,}464 narrated videos) and the unique-video pool of SurgVLM-DB and Surg$\Sigma$-DB ($\sim$1{,}590)~\cite{SurgVLM,SurgSigma}.

\paragraph{Diversity.}
The corpus spans $S = 18$ surgical specialties and over $5000$ procedure types canonicalized against CPT codes. Surgery-type composition is heterogeneous: $6{,}182$ open-surgery videos ($40.4\%$), $3{,}892$ laparoscopic ($25.5\%$), $3{,}827$ robotic ($25.0\%$), $647$ endoscopic ($4.2\%$), and $743$ videos with other modalities including cases combining surgery types, microsurgical and arthroscopic procedures ($4.9\%$) (Figure~\ref{fig:specialty_distribution}).
 SurgAtlas is the first public surgical video--language dataset to include open surgery at scale; all prior public resources are restricted to scoped procedures (Figure~\ref{fig:scale}). 

\paragraph{Richness.}
Each video is annotated at up to four levels of granularity, yielding $300K$ segment-level captions, $81K$ step- and phase-level descriptions, $12.2K$ video-level summaries, and $400K$ reasoning question--answer pairs. The reasoning annotations are organized hierarchically across five broad and ten fine-grained categories, supporting both fine-grained supervision during training and structured analysis at multiple granularities. In contrast to prior surgical VLP datasets that provide a single annotation level (e.g., clip captions in~\cite{SurgVLP,SurgLaVi}) or a single QA template (e.g.,~\cite{SUREON}), SurgAtlas is, to our knowledge, the only public surgical video--language dataset providing all four levels jointly.

\paragraph{Quality.}
We characterize annotation quality at two levels. \emph{Source grounding:} segment- and step-level annotations originate from clinically authored sources---surgeon narration in Tier~1 and surgeon-authored on-screen procedural text in Tier~2---and pass through the LLM refinement operators $\mathcal{R}_{\text{LLM}}$ and $\mathcal{R}_{\text{LLM}}^{\text{ocr}}$ only to correct upstream transcription errors introduced by Whisper and OCR, normalize surgical terminology, and remove non-procedural content, never to invent surgical content. Quality assurance therefore concentrates on the QA layer, where LLM operators contribute new content. \emph{Automated:} across all candidates produced in Stage~5, the verifier $\mathcal{R}_{\text{judge}}$ rejects $7\%$. \emph{Human:} on the expert-validated subset $\mathcal{A}^{\text{exp}}$ (Section~\ref{sec:expert_subset}), 2{,}960 candidate question--answer pairs were independently reviewed by two surgical residents, of which 2{,}890 were unanimously accepted (97.6\%) and 70 were rejected by at least one reviewer. Raw agreement on the binary accept decision is 98.2\% across both partitions. The accepted subset $\mathcal{A}^{\text{exp}}$ comprises 1{,}462 pairs from open-surgery videos and 1{,}428 pairs from minimally invasive videos. 

\section{Experiments}
\label{sec:experiment}

\subsection{SurgAtlas-VLM: Aligning Qwen3-VL with SurgAtlas}
\label{sec:surgatlas_vlm}

To exploit the multigranular supervision in $\mathcal{D}$, we propose SurgAtlas-VLM, built on Qwen3-VL-8B~\cite{Qwen3VL}. Training proceeds in two stages: a captioning pretraining stage that aligns the visual encoder, projector, and language model to surgical content, followed by an instruction tuning stage that adds reasoning and structured-task supervision. 

\paragraph{Stage 1: Captioning pretraining.}
Captioning pretraining itself proceeds in three steps to progressively unfreeze model components and align them to the surgical domain. \emph{Step~1 (projector only).} We train only the MLP projector with the vision encoder and LLM frozen on $\mathcal{D}_{\text{cap}} = \bigcup_i (\mathcal{A}_i^{\text{seg}} \cup \mathcal{A}_i^{\text{step}} \cup \mathcal{A}_i^{\text{vid}})$ at learning rate $\eta_1 = 10^{-4}$, bringing the visual features into the surgical language space. \emph{Step~2 (projector + LLM).} We jointly fine-tune the projector and LLM (vision frozen) on $\mathcal{D}_{\text{cap}}$ at $\eta_2 = 2 \times 10^{-5}$, teaching the LLM surgical vocabulary and descriptive ability. \emph{Step~3 (full model).} We finally unfreeze the vision encoder and train all components on a curated subset $\mathcal{D}_{\text{cap}}^{\text{clean}} \subset \mathcal{D}_{\text{cap}}$ that excludes raw narration captions and retains only step-level descriptions, video summaries, and high-confidence segment captions, at $\eta_3 = 5 \times 10^{-6}$ for the LLM/projector and $\eta_3^{v} = 2 \times 10^{-6}$ for the vision encoder, for one epoch. This step adapts the visual representations to the surgical visual domain---particularly the underrepresented open-surgery regime---without overfitting to noisy ASR-derived supervision.

\paragraph{Stage 2: Instruction tuning.}
We then train the projector and LLM (vision frozen, having been adapted in Stage~1 Step~3) on $\mathcal{D}_{\text{inst}} = \bigcup_i \mathcal{A}_i^{\text{qa}} \cup \mathcal{D}^{\text{IT}}$ at learning rate $5 \times 10^{-6}$ for one epoch, where $\mathcal{D}^{\text{IT}}$ is the converted public-dataset supervision from Stage~7. This stage teaches the model to answer structured questions and to perform reasoning grounded in the operative scene.

\subsection{Results}
\label{sec:benchmarks}
\subsubsection{Standard Surgical Benchmarks}

We evaluate SurgAtlas-VLM on three classes of established benchmarks: phase and action recognition (Cholec80, HeiChole, MultiBypass140), instrument--verb--target triplet recognition (CholecT50), and critical view of safety assessment (Endoscapes-CVS). Since neither SureonVLM~\cite{SUREON} nor SurgVLM~\cite{SurgVLM} weights are publicly available at the time of writing, we report each baseline's numbers as published in its original paper and follow the metric convention of the source paper for each table. We omit contrastive surgical VLP baselines because their similarity-based prediction protocol is structurally distinct from generative inference. All evaluations sample at $1$~fpm on Cholec80/HeiChole and $1$~fp$3$m on MultiBypass140 following~\cite{SUREON,SurgVLM}.

\paragraph{Phase and action recognition.}

\begin{table}[h!]
\centering
\caption{Macro F1 (\%) on phase and action recognition benchmarks. Baselines as reported in ~\cite{SUREON}. Best results are shown in bold.}
\label{tab:phase_macro}
\setlength{\tabcolsep}{4pt}
\renewcommand{\arraystretch}{1.05}
\small

\begin{tabular}{@{}l S S S S@{}}
\toprule
Method
& {Cholec80}
& {\makecell{HeiChole\\Phase}}
& {\makecell{HeiChole\\Action}}
& {\makecell{Multi-\\Bypass140}} \\
\midrule
GPT-5.1        & 36.0 & 29.0 & 18.0 & 13.0 \\
Gemini 3.1 Pro & 47.0 & 35.0 & 21.0 & 22.0 \\
Qwen3-VL-8B    & 17.0 & 12.0 & 17.0 &  8.0 \\
SureonVLM      & 63.0 & 41.0 &  4.0 & {\bfseries 40.0} \\
\midrule
\textbf{SurgAtlas-VLM (ours)}
& {\bfseries 68.2}
& {\bfseries 54.5}
& {\bfseries 55.8}
& {\bfseries 40.0} \\
\bottomrule
\end{tabular}
\end{table}

Table~\ref{tab:phase_macro} reports macro F1, as used by ~\cite{SUREON} on the four benchmarks where it reports results: phase recognition on Cholec80, HeiChole, and MultiBypass140, and action recognition on HeiChole. SurgAtlas-VLM (8B) outperforms SureonVLM (also 8B) by $+5.2$ on Cholec80 phase, $+13.5$ on HeiChole phase, significantly outperforms it on HeiChole action, and matches it on MultiBypass140. Frontier general-domain VLMs trail substantially despite operating at much larger scale. Table~\ref{tab:detailed_results} reports detailed Cholec80 metrics following SurgVLM~\cite{SurgVLM}'s reporting convention against their broader baseline set; SurgAtlas-VLM achieves the highest precision across all evaluated models and approaches SurgVLM-72B's accuracy at $\sim 9\times$ fewer parameters.

\paragraph{Triplet recognition and CVS assessment.}

Table~\ref{tab:detailed_results} also reports CholecT50 triplet mAP and Endoscapes-CVS criterion accuracy respectively, both following~\cite{SurgVLM}'s convention. SurgAtlas-VLM achieves $5.5$ triplet mAP, exceeding SurgVLM-72B ($4.8$) and substantially exceeding all general-domain baselines (best: GPT-4o at $2.6$). On CVS, we average $77.7\%$ accuracy across the three criteria, outperforming both SurgVLM-7B ($76.9\%$) and SurgVLM-72B ($76.6\%$) while substantially outperforming general-domain VLMs. 

\label{sec:exp_standard}
\begin{table}[h!]
\centering
\scriptsize
\caption{Detailed surgical video understanding benchmarks. Cholec80 reports phase recognition metrics, CholecT50 reports triplet recognition mAP, and Endoscapes-CVS reports accuracy for the three critical-view-of-safety criteria. 
Avg.\ is the mean across the three CVS criteria. All values are percentages.}
\label{tab:detailed_results}
\setlength{\tabcolsep}{3.5pt}
\renewcommand{\arraystretch}{1.08}

\begin{tabular}{@{}l
                S S S
                S
                S S S S@{}}
\toprule
& \multicolumn{3}{c}{Cholec80}
& \multicolumn{1}{c}{CholecT50}
& \multicolumn{4}{c}{Endoscapes-CVS} \\
\cmidrule(lr){2-4}
\cmidrule(lr){5-5}
\cmidrule(lr){6-9}
Metric
& {Acc.}
& {Recall}
& {Prec.}
& {Triplet mAP}
& {Avg.}
& {C1}
& {C2}
& {C3} \\
\midrule
InternVL3-8B
& 23.9 & 15.2 & 15.2
& 2.4
& 48.2 & 40.0 & 53.3 & 51.4 \\

Qwen2.5-VL-7B
& 30.5 & 16.7 & 26.2
& 2.4
& 65.9 & 56.1 & 82.4 & 59.2 \\

GPT-4o
& 36.4 & 31.0 & 33.0
& 2.6
& 6.7 & 6.7 & 5.9 & 7.5 \\

Gemini 2.0 Flash
& 38.9 & 36.8 & 40.0
& 2.5
& 59.6 & 47.8 & 63.9 & 67.1 \\

SurgVLM-7B
& 70.3 & 61.9 & 59.8
& 2.4
& 76.9 & 75.3 & 82.4 & 72.9 \\

SurgVLM-72B
& {\bfseries 76.4}
& {\bfseries 70.8}
& 66.0
& 4.8
& 76.6
& 76.1
& {\bfseries 83.1}
& 70.6 \\
\midrule
\textbf{SurgAtlas-VLM (ours)}
& 66.9
& 66.8
& {\bfseries 73.5}
& {\bfseries 5.5}
& {\bfseries 77.7}
& {\bfseries 77.1}
& 82.9
& {\bfseries 73.0} \\
\bottomrule
\end{tabular}
\end{table}

\begin{table*}[t!]
\centering
\caption{VQA results on the expert-validated subset $\mathcal{A}^{\text{exp}}$ for both partitions: minimally invasive (MIS, 1,428 pairs) and open surgery (Open, 1,462 pairs). LLM-judge accuracy (\%) reported per reasoning category and overall. For GPT and Gemini models, we sample 4 frames per QA clip. Best per column in bold.}
\label{tab:vqa_expert}
\small
\setlength{\tabcolsep}{4pt}
\renewcommand{\arraystretch}{1.1}
\begin{tabular}{l cc cc cc cc cc cc cc}
\toprule
& \multicolumn{2}{c}{Perception \& ID}
& \multicolumn{2}{c}{Action / state}
& \multicolumn{2}{c}{Operative reas.}
& \multicolumn{2}{c}{Temporal / pred.}
& \multicolumn{2}{c}{Risk anatomy ID}
& \multicolumn{2}{c}{Overall} \\
\cmidrule(lr){2-3} \cmidrule(lr){4-5} \cmidrule(lr){6-7} \cmidrule(lr){8-9} \cmidrule(lr){10-11} \cmidrule(lr){12-13}
Method & MIS & Open & MIS & Open & MIS & Open & MIS & Open & MIS & Open & MIS & Open \\
\midrule
\multicolumn{13}{l}{\textit{General-domain VLMs}} \\
GPT-5~\cite{GPT5}            & 34.0 & 36.5 & 39.9 & 32.5 & 49.2 & \textbf{51.2} & 38.7 & 34.8 & 45.1 & 67.2 & 39.2 & 37.6 \\
GPT-5.1~\cite{GPT5}          & 34.2 & 35.2 & 41.7 & 34.3 & \textbf{49.7} & 47.6 & 40.8 & 32.9 & 51.0 & 59.0 & 40.4 & 37.0 \\
Gemini 2.5 Pro~\cite{comanici2025gemini25pushingfrontier} & 28.7 & 27.2 & 35.9 & 29.2 & 38.7 & 32.9 & 34.8 & 31.2 & 46.0 & 56.5 & 34.1 & 30.3 \\
Qwen3-VL-8B~\cite{Qwen3VL}   & 24.3 & 16.3 & 19.9 & 19.4 & 21.8 & 18.1 & 33.8 & 22.6 & 47.1 & 50.8 & 24.0 & 19.9 \\
Qwen3-VL-32B~\cite{Qwen3VL}  & 30.1 & 21.9 & 24.2 & 24.6 & 28.5 & 31.3 & 33.8 & 27.1 & 47.1 & 62.3 & 28.6 & 26.3 \\
\midrule
\multicolumn{13}{l}{\textit{Ours (Qwen3-VL-8B fine-tuned on SurgAtlas)}} \\
SurgAtlas-VLM (SurgAtlas)    & \textbf{42.3} & \textbf{41.0} & 42.5 & 34.5 & 37.3 & 37.9 & 44.4 & 40.0 & \textbf{58.8} & 67.2 & 42.5 & 39.0 \\
SurgAtlas-VLM (\,+ public)   & 41.6 & 39.4 & \textbf{42.7} & \textbf{36.0} & 39.9 & 38.5 & \textbf{46.5} & \textbf{41.3} & \textbf{58.8} & \textbf{68.9} & \textbf{42.9} & \textbf{39.3} \\
\bottomrule
\end{tabular}
\end{table*}

\subsubsection{SurgAtlas benchmarks.}
Table~\ref{tab:vqa_expert} reports LLM-judge accuracy on $\mathcal{A}^{\text{exp}}$ by broad reasoning category and overall, evaluated separately on the open and MIS partitions. We report two variants of our model. \emph{SurgAtlas only} uses Stage~2 instruction tuning on the SurgAtlas reasoning QA pairs $\mathcal{A}^{\text{qa}}$ alone, while \emph{SurgAtlas + public} additionally incorporates $\mathcal{D}^{\text{IT}}$---the public-dataset supervision obtained by VQA-converting the training splits of 18 public surgical datasets. Both variants share the same Qwen3-VL-8B backbone and Stage~1 captioning recipe. The full model outperforms every general-domain baseline on both partitions, overall MIS (42.9 vs.\ GPT-5.1 at 40.4) and overall open surgery (39.3 vs.\ GPT-5 at 37.6), despite using $\sim 10$--$30\times$ fewer parameters. The SurgAtlas-only variant performs comparably overall, indicating that the SurgAtlas corpus drives the bulk of the improvement; $\mathcal{D}^{\text{IT}}$ contributes incremental gains on the action and temporal categories where converted public benchmarks provide aligned supervision, while perception-oriented categories favor the SurgAtlas-only variant. SurgAtlas variants lead on every broad category except operative reasoning, where the GPT-5 family leads by $\sim 10$ points on both partitions---reflecting that deep clinical justification (why a maneuver is performed) still benefits from frontier-scale pretraining and broad world knowledge in ways that an 8B surgical model does not yet replicate.

\paragraph{Open versus minimally invasive surgery.}
Every general-domain baseline performs worse on open surgery than on MIS---GPT-5.1 by 3.4 points overall, Gemini 2.5 Pro by 3.8, Qwen3-VL-8B by 4.1, Qwen3-VL-32B by 2.3---with the gap widening for models with less surgical exposure during pretraining. Both SurgAtlas variants exhibit a similar partition gap ($-3.5$ to $-3.6$), but absolute open-surgery performance (39.3) exceeds the MIS performance of every general-domain baseline except GPT-5.1. To our knowledge, this is the first systematic VLM evaluation on open-surgery reasoning, made possible by SurgAtlas's explicit open-surgery component.

\subsection{Ablation}

\paragraph{Cross-regime training ablation.}
We isolate the contribution of each regime by training SurgAtlas on (a) the MIS partition only, (b) the open partition only, and (c) the combined corpus, evaluating on $\mathcal{A}^{\text{exp}}$ partitioned by surgical regime and on EgoSurgery-Phase~\cite{EgoSurgery} as an out-of-domain probe (Table~\ref{tab:cross_regime_ablation}). 

EgoSurgery-Phase is a real egocentric open-surgery dataset captured with cameras attached to the surgeon's head, spanning multiple surgical types and phase labels. Each frame captures not just the surgical action but the surrounding operating-room context---interactions among surgeons, assistant surgeons, anesthesiologists, perfusionists, and nurses, alongside varied operative settings and lighting conditions. It therefore provides a demanding test of whether learned representations generalize from canonical surgical-field views that predominantly depict surgical actions to wearable open-surgery recordings that depict the broader operating-room scene.
\emph{In-domain}, the combined model is best on both partitions of $\mathcal{A}^{\text{exp}}$ ($42.5\%$ MIS, $39.0\%$ Open). Training on only one regime reduces accuracy on the other by $\sim 10$ points: SurgAtlas-MIS reaches only $29.1\%$ on Open, while SurgAtlas-Open reaches only $32.2\%$ on MIS. This indicates poor cross-regime generalization in both directions. \emph{Out-of-domain}, the open-only variant takes the lead. On EgoSurgery-Phase, SurgAtlas-Open reaches $62.5\%$, surpassing both the combined variant ($56.1\%$) and GPT-5.1 ($55.9\%$). SurgAtlas-MIS, consistent with its in-domain behavior on open data, transfers poorly: it drops to $24.8\%$, nearly $20$ points below the Qwen3-VL~$8$B base ($47.8\%$). This gap reflects the visual content of the training data: open-surgery videos expose the model to varied camera angles, lighting, personnel, and operative context that overlap with egocentric open surgery footage, whereas minimally invasive clips remain confined to a narrow surgical-field view with little transferable scene content. 
Together, these ablations demonstrate that the visual distribution of open surgery differs from that of minimally invasive surgery, reinforcing the value of establishing it as a separate benchmark regime. They also highlight its relevance for real-world operating-room deployment, where cameras may capture broader operative context rather than only the surgical field.
\begin{table}[h]
\centering
\caption{Accuracy (\%) on the expert-validated subset $\mathcal{A}^{\text{exp}}$ partitioned by surgical regime (LLM-judge), and zero-shot on EgoSurgery~\citep{EgoSurgery}, a held-out open-surgery phase benchmark with a distinct egocentric wide-angle viewpoint. SurgAtlas variants are trained on the MIS partition only, the open partition only, or the combined corpus. Best per column in bold.}
\label{tab:cross_regime_ablation}
\small
\setlength{\tabcolsep}{6pt}
\renewcommand{\arraystretch}{1.05}
\begin{tabular}{l cc c}
\toprule
& \multicolumn{2}{c}{$\mathcal{A}^{\text{exp}}$} & EgoSurgery \\
\cmidrule(lr){2-3} \cmidrule(lr){4-4}
Model & MIS & Open & (zero-shot) \\
\midrule
Qwen3-VL~$8$B    & 24.0 & 19.9 & 43.8 \\
GPT-5.1                     & 40.4 & 37.0 & 55.9 \\
\midrule
SurgAtlas-VLM (MIS only)        & 40.1 & 29.1 & 24.8 \\
SurgAtlas-VLM (Open only)       & 32.2 & 38.9 & \textbf{62.5} \\
SurgAtlas-VLM (combined)        & \textbf{42.5} & \textbf{39.0} & 56.1 \\
\bottomrule
\end{tabular}
\end{table}
\section{Conclusion}
We introduce SurgAtlas, a surgical video--language dataset of $15{,}291$ videos ($2{,}391$ hours) across $18$ specialties and both open and minimally invasive surgery. Its newly collected public corpus is paired with $793$K multi-granular annotations spanning captions, procedural descriptions, video summaries, and hierarchical reasoning QA. SurgAtlas further establishes an expert-validated benchmark across both open and MIS surgical regimes enabling systematic evaluation. By making open surgery a first-class setting, SurgAtlas broadens the scope of surgical video--language learning and evaluation across diverse operative contexts. We release SurgAtlas to support surgical multimodal models that generalize across procedures and specialties, advancing applications from surgical training to intraoperative decision support.

\begin{acks}
    This research was funded, in part, by the U.S. Government, under Agreement No. 1AY2AX000062. The views and conclusions contained in this document are those of the authors and should not be interpreted as representing the official policies, either expressed or implied, of the U.S. Government. This work was also supported by The American Board of Thoracic Surgery and by the National Heart, Lung, and Blood Institute of the National Institutes of Health under Award Number R01HL146619. The content is solely the responsibility of the authors and does not necessarily represent the official views of the National Institutes of Health.
\end{acks}

\bibliographystyle{ACM-Reference-Format}
\bibliography{sample-base}

\clearpage
\appendix
\label{app:specialty_distribution}
\twocolumn[{%
\begin{minipage}{\textwidth}

\section{Supplementary material}
\subsection{Specialty distribution.}

\centering
\includegraphics[width=\textwidth]
    {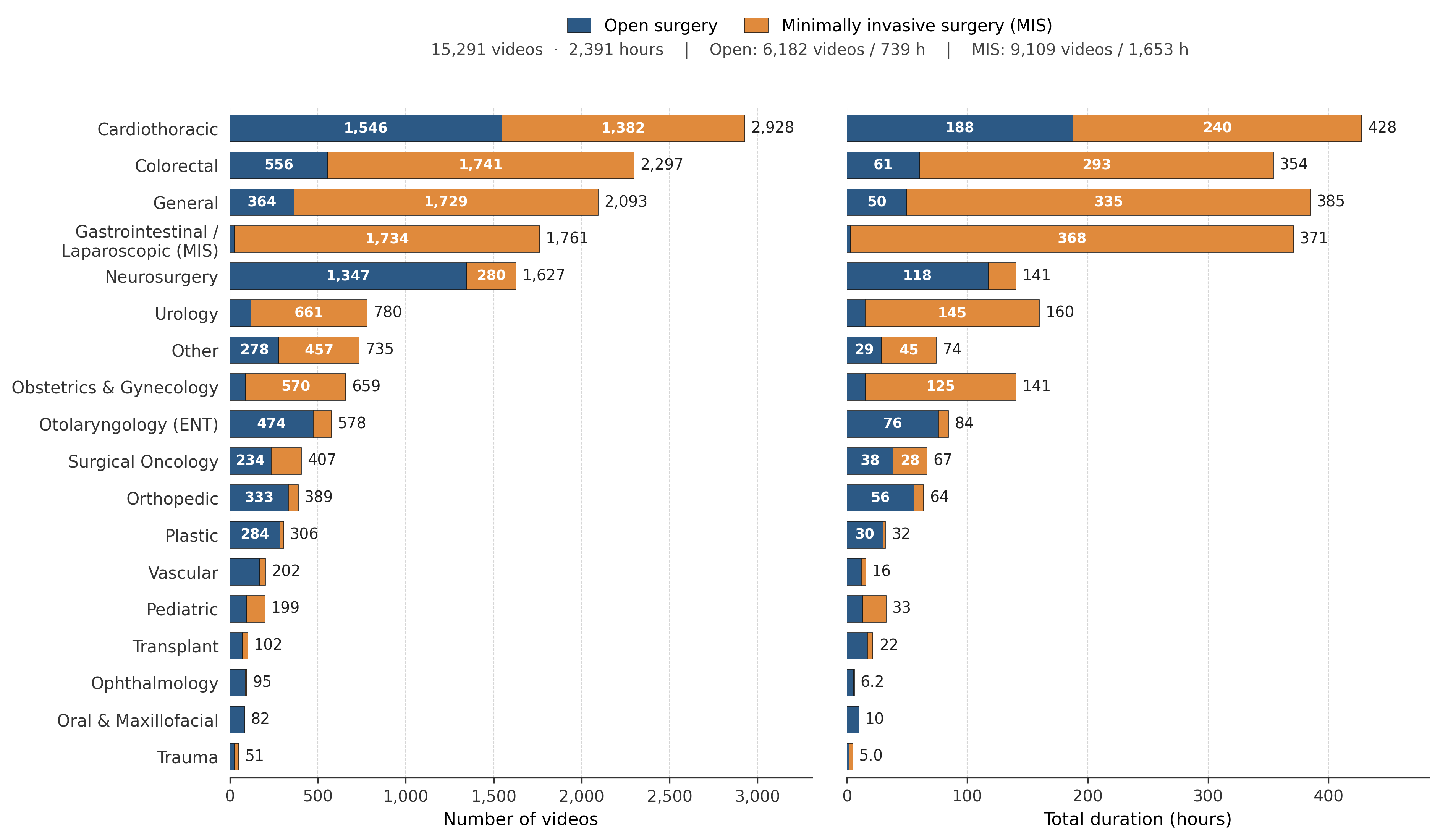}

\captionof{figure}{
\textbf{Per-specialty composition of SurgAtlas.}
Number of videos (left) and total duration in hours (right) across
the $18$ surgical specialties, decomposed into open surgery (blue)
and minimally invasive surgery (orange). Specialties are sorted by
total video count. Five specialties---Cardiothoracic, Colorectal,
General, Gastrointestinal/Laparoscopic, and Neurosurgery---account
for $\sim 70\%$ of the corpus, while the open/MIS balance within each
specialty closely tracks clinical practice.
}
\label{fig:specialty_distribution}

\end{minipage}
}]
\noindent


Figure~\ref{fig:specialty_distribution} reports the per-specialty composition of SurgAtlas across the $18$ surgical specialties, decomposed into open and minimally invasive (MIS) components. The full corpus comprises $15{,}291$ videos and $2{,}391$ hours of footage, split into $6{,}182$ open-surgery videos ($739$ h) and $9{,}109$ MIS videos ($1{,}653$ h). The top five specialties by video count---Cardiothoracic ($2{,}928$), Colorectal ($2{,}297$), General ($2{,}093$), Gastrointestinal/Laparoscopic ($1{,}761$), and Neurosurgery ($1{,}627$)---together account for $\sim 70\%$ of the corpus. The open/MIS balance within each specialty mirrors clinical practice: neurosurgical, orthopedic, plastic, vascular, transplant, ophthalmic, and oral-maxillofacial procedures are overwhelmingly open, while gastrointestinal/laparoscopic, colorectal, general, urological, and obstetric/gynecological videos are dominated by minimally invasive techniques. Cardiothoracic is the most balanced specialty ($1{,}546$ open vs.\ $1{,}382$ MIS), reflecting the coexistence of conventional cardiac surgery and thoracoscopic approaches at scale.

\subsection{VQA taxonomy examples}
\label{app:vqa_examples}

\begin{figure*}[t]
    \centering
    \includegraphics[width=\textwidth]{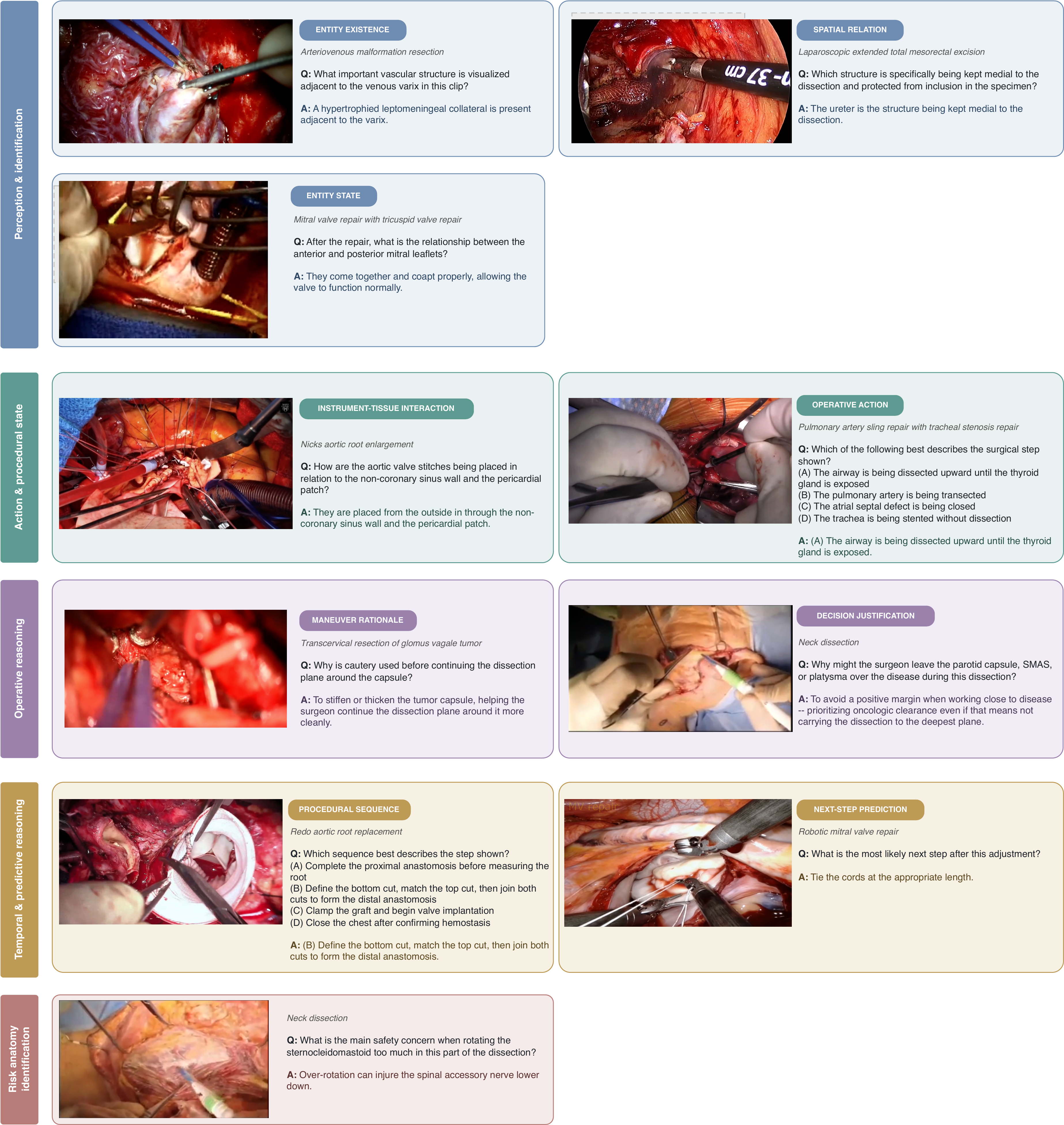}
\caption{\textbf{Representative VQA examples across the SurgAtlas taxonomy.} One expert-validated example per fine-grained category, spanning open and minimally invasive procedures in both open-ended and multiple-choice formats.}
    \label{fig:vqa_examples}
\end{figure*}

Figure~\ref{fig:vqa_examples} presents representative visual question--answer pairs spanning all ten fine-grained categories of the SurgAtlas taxonomy, drawn from distinct source videos across both open and minimally invasive procedures. Each example pairs a short video clip with a question targeting a specific reasoning skill---from perceptual grounding (entity existence, spatial relation) through action understanding, operative reasoning, and temporal prediction---and includes both open-ended and multiple-choice formats. All shown pairs received maximum scores for groundedness, temporal alignment, answerability, and non-triviality, with low hallucination risk.

\newpage
\null
\newpage

\subsection{Procedure-Level Construction Examples}
\label{app:construction_examples}

This section illustrates the two complementary paths by which SurgAtlas produces step- and phase-level supervision: procedure-conditioned expansion of on-screen phase labels (Tier~2) and temporal aggregation of narrated segments (Tier~1).

\subsubsection{OCR-derived supervision and phase fallback (Tier~2)}

For Tier~2 videos, each cleaned on-screen phase label is aligned to a procedure-conditioned canonical step list. Each knowledge-base entry pairs a canonical step name with a procedure-specific description. When the alignment score exceeds the acceptance threshold $\theta_s$, the matched step's description replaces the shorter OCR label as the training target; otherwise the normalized OCR label is retained unchanged as a \emph{phase-fallback} target. Of the $24{,}669$ Tier~2 annotations used for training, $8{,}671$ ($35.1\%$) received knowledge-augmented descriptions and $15{,}998$ ($64.9\%$) used phase fallback. This conservative policy prevents an uncertain canonical-step match from injecting procedural detail beyond what the source on-screen annotation supports.

The examples below show, for each case, the source OCR phase, the matched canonical step, and the resulting training target. To illustrate the structure of a procedure-conditioned step list, the complete knowledge base for Example~1 is shown; Examples~2 and~3 report only the matched step for brevity.

\paragraph{Example 1 --- Critical view of safety (laparoscopic cholecystectomy).}
\emph{Source phase:} ``Critical view of safety and cystic duct clipping'' $\rightarrow$ \emph{matched canonical step:} ``Critical view of safety.''

The complete procedure-conditioned step-list knowledge base generated
for this example is shown below:
\begin{tcolorbox}[
  breakable,
  colback=gray!3,
  colframe=gray!40,
  title=\textbf{Procedure-conditioned canonical step list: Laparoscopic cholecystectomy},
  fonttitle=\small,
  fontupper=\footnotesize,
  left=2mm, right=2mm, top=1mm, bottom=1mm
]
Generated by \texttt{gpt-5.4-mini}, conditioned on the procedure label. Used only as an expansion target; not an expert-validated clinical guideline.
\begin{enumerate}[leftmargin=*, itemsep=1pt, topsep=2pt]
  \item \textbf{Abdominal access and pneumoperitoneum.} Establish laparoscopic access with a Veress needle or open technique and insufflate with carbon dioxide. Place camera and working trocars under direct vision.
  \item \textbf{Initial laparoscopic inspection.} Survey the abdominal cavity and inspect the gallbladder, liver, and surrounding structures; identify adhesions or inflammation.
  \item \textbf{Gallbladder retraction and exposure.} Grasp the fundus and infundibulum to retract cephalad and laterally, placing the hepatocystic triangle under tension.
  \item \textbf{Dissection of Calot's triangle.} Divide peritoneal coverings and fibroareolar tissue around the cystic duct and artery; clear the lower third of the gallbladder from the cystic plate.
  \item \textbf{Critical view of safety.} \emph{(matched step)} Ensure only two structures enter the gallbladder and the hepatocystic triangle is cleared; confirm the lower gallbladder is separated from the liver bed before dividing any tubular structures.
  \item \textbf{Clip and divide the cystic duct.} Apply clips with proximal and distal control; divide between clips.
  \item \textbf{Clip and divide the cystic artery.} Identify the cystic artery clearly, clip proximally and distally, and divide between clips.
  \item \textbf{Gallbladder dissection from the liver bed.} Separate the gallbladder from the cystic plate with an energy device, maintaining hemostasis and avoiding bile spillage.
  \item \textbf{Specimen retrieval.} Place the gallbladder in a retrieval bag and remove through a port site, enlarging the fascial opening if needed.
  \item \textbf{Hemostasis, irrigation, and bile-leak check.} Inspect the fossa and duct/artery stumps for bleeding or bile leak; irrigate and aspirate until clean.
  \item \textbf{Port removal and closure.} Release pneumoperitoneum under vision, remove trocars, and close fascial defects and skin.
\end{enumerate}
\emph{Source OCR phase} ``Critical view of safety and cystic duct clipping'' aligned to step~5, whose description became the training target.
\end{tcolorbox}

\paragraph{Example 2 --- Level II dissection (supraomohyoid neck dissection).}
\emph{Source phase:} ``Dissecting the spinal accessory nerve and level IIb'' $\rightarrow$ \emph{matched canonical step:} ``Level II dissection.''\\
\textbf{Q:} Describe the surgical step shown.\\
\textbf{A:} Lymphatic tissue along the upper internal jugular chain is mobilized and removed, taking care to identify and preserve the spinal accessory nerve, hypoglossal nerve, and internal jugular vein. The specimen is dissected from the upper neck boundaries.

\paragraph{Example 3 --- Semi-skeletonized dissection (internal thoracic artery harvest).}
\emph{Source phase:} ``Separation of IMA and satellite veins from fascia'' $\rightarrow$ \emph{matched canonical step:} ``Semi-skeletonized dissection of the artery.''\\
\textbf{Q:} Describe the surgical step shown.\\
\textbf{A:} The artery is freed from surrounding veins, fascia, and adipose tissue while leaving a thin cuff of tissue on one side of the vessel. The dissection proceeds along the desired length of the conduit with care to avoid thermal or traction injury.

\begin{table*}[!t]
\centering
\caption{VQA results on the expert-validated subset $\mathcal{A}^{\text{exp}}$ broken down by all 10 fine-grained reasoning categories. LLM-judge accuracy (\%) reported per partition. Best result per category and partition in bold.}
\label{tab:vqa_finegrained_appendix}
\setlength{\tabcolsep}{2pt}
\renewcommand{\arraystretch}{0.95}
\begin{tabular}{l cc cc cc cc cc cc cc}
\toprule
& \multicolumn{10}{c}{\textit{General-domain VLMs}}
& \multicolumn{4}{c}{\textit{Surgical VLMs (ours)}} \\
\cmidrule(lr){2-11} \cmidrule(lr){12-15}
& \multicolumn{2}{c}{GPT-5}
& \multicolumn{2}{c}{GPT-5.1}
& \multicolumn{2}{c}{\makecell{Gemini\\2.5 Pro}}
& \multicolumn{2}{c}{\makecell{Qwen3-VL\\8B}}
& \multicolumn{2}{c}{\makecell{Qwen3-VL\\32B}}
& \multicolumn{2}{c}{\makecell{SurgAtlas\\only}}
& \multicolumn{2}{c}{\makecell{SurgAtlas\\+ public}} \\
\cmidrule(lr){2-3} \cmidrule(lr){4-5} \cmidrule(lr){6-7} \cmidrule(lr){8-9} \cmidrule(lr){10-11} \cmidrule(lr){12-13} \cmidrule(lr){14-15}
Category & MIS & Open & MIS & Open & MIS & Open & MIS & Open & MIS & Open & MIS & Open & MIS & Open \\
\midrule
\multicolumn{15}{l}{\textit{Perception \& identification}} \\
\quad Entity existence
& 36.5 & 36.9 & 34.5 & 35.9 & 29.7 & 28.9 & 26.5 & 18.5 & 33.0 & 20.0 & \textbf{46.0} & \textbf{47.7} & 45.5 & 47.2 \\
\quad Entity state
& \textbf{47.1} & 36.0 & 43.5 & 38.0 & 35.3 & 25.7 & 29.1 & 18.0 & 34.9 & 27.0 & 43.0 & \textbf{41.0} & 44.2 & 33.0 \\
\quad Spatial relation
& 25.8 & \textbf{36.3} & 29.8 & 33.2 & 24.9 & 26.2 & 20.1 & 13.2 & 25.1 & 21.1 & \textbf{38.2} & 34.2 & 36.7 & 34.7 \\
\midrule
\multicolumn{15}{l}{\textit{Action \& procedural state}} \\
\quad \makecell[l]{Instrument--tissue\\interaction}
& \textbf{50.4} & 42.2 & 49.6 & 44.8 & 40.0 & 41.0 & 24.3 & 25.0 & 27.0 & 29.3 & 44.3 & \textbf{45.7} & 43.5 & \textbf{45.7} \\
\quad Operative action
& 37.1 & 30.1 & 39.6 & 31.8 & 34.8 & 26.3 & 18.8 & 18.1 & 23.5 & 23.5 & 42.1 & 31.8 & \textbf{42.5} & \textbf{33.7} \\
\midrule
\multicolumn{15}{l}{\textit{Operative reasoning}} \\
\quad Maneuver rationale
& \textbf{58.0} & \textbf{49.2} & 53.6 & 42.9 & 42.0 & 33.3 & 27.5 & 12.7 & 31.9 & 28.6 & 33.3 & 33.3 & 37.7 & 28.6 \\
\quad Decision justification
& 44.3 & \textbf{52.4} & \textbf{47.5} & 50.5 & 36.9 & 32.7 & 18.5 & 21.4 & 26.6 & 33.0 & 39.5 & 40.8 & 41.1 & 44.7 \\
\midrule
\multicolumn{15}{l}{\textit{Temporal \& predictive reasoning}} \\
\quad Procedural sequence
& 41.4 & 37.0 & 44.0 & 34.5 & 35.7 & 35.3 & 35.3 & 23.5 & 33.6 & 30.3 & 46.6 & 41.2 & \textbf{48.3} & \textbf{45.4} \\
\quad Next-step prediction
& 26.9 & 27.8 & 26.9 & 27.8 & 30.8 & 18.4 & 26.9 & 19.4 & 34.6 & 16.7 & 34.6 & \textbf{36.1} & \textbf{38.5} & 27.8 \\
\midrule
\multicolumn{15}{l}{\textit{Risk anatomy identification}} \\
\quad \makecell[l]{Risk anatomy\\identification}
& 45.1 & 67.2 & 51.0 & 59.0 & 46.0 & 56.5 & 47.1 & 50.8 & 47.1 & 62.3 & \textbf{58.8} & 67.2 & \textbf{58.8} & \textbf{68.9} \\
\midrule
\textbf{Overall}
& 39.2 & 37.6 & 40.4 & 37.0 & 34.1 & 30.3 & 24.0 & 19.9 & 28.6 & 26.3 & 42.5 & 39.0 & \textbf{42.9} & \textbf{39.3} \\
\bottomrule
\end{tabular}
\end{table*}

\subsubsection{Narration-derived step supervision (Tier~1)}

For Tier~1 videos, temporally adjacent refined narration segments that share a common procedural objective are grouped into step-level intervals. The grouping model receives the ordered segment descriptions, their timestamps, and video-level procedure metadata, and emits a concise phase name together with a unified step description. Unlike the Tier~2 path, this process does not consult the procedure-conditioned knowledge base; the step description is synthesized directly from the contributing narration. The two examples below were chosen because every clause in the final description is directly supported by the source narration.

\paragraph{Example 4 --- Inferior mesenteric artery ligation (robotic TME).}
\emph{Interval:} $143.7$--$182.1$\,s. The contributing narration described two consecutive actions: ICG near-infrared imaging to identify the left colic artery and select a division point near the IMA origin, followed by low ligation of the IMA under white light while preserving the left colic branch.\\
\textbf{Q:} Describe the surgical step being performed in this clip.\\
\textbf{A:} After dissection around the inferior mesenteric artery root, ICG fluorescence was used to identify the left colic artery and guide low ligation of the inferior mesenteric artery while preserving the left colic branch.

\paragraph{Example 5 --- Nerve-stimulator-guided pudendal block.}
\emph{Interval:} $49.2$--$96.6$\,s. The contributing narration described connecting a nerve-stimulator needle, applying stimulating current, verifying pudendal-nerve stimulation via ipsilateral anal-sphincter contraction, and adjusting stimulation to confirm the needle was not intraneural.\\
\textbf{Q:} Describe the surgical step being performed in this clip.\\
\textbf{A:} A nerve-stimulator needle is connected and used to identify the pudendal nerve by eliciting ipsilateral anal-sphincter contraction, while adjusting stimulation to confirm safe needle position and avoid intraneural placement.

\noindent Together, these examples illustrate the two complementary construction paths: conservative procedure-conditioned expansion of on-screen phase labels in Tier~2, and temporal aggregation of locally grounded narration in Tier~1.
\begin{table*}[!t]
\centering
\caption{VQA results on the held-out test split of $\mathcal{A}^{\text{qa}}$ ($\sim 5\%$  held out from training) broken down by all 10 fine-grained reasoning categories. LLM-judge accuracy (\%) reported per partition. This split is independent of the expert-validated subset $\mathcal{A}^{\text{exp}}$ reported in Table~\ref{tab:vqa_expert}. Best result per category and partition in bold.}
\label{tab:vqa_finegrained_testset}
\setlength{\tabcolsep}{3pt}
\renewcommand{\arraystretch}{0.95}
\begin{tabular}{l cc cc cc cc cc}
\toprule
& \multicolumn{6}{c}{\textit{General-domain VLMs}}
& \multicolumn{4}{c}{\textit{Surgical VLMs (ours)}} \\
\cmidrule(lr){2-7} \cmidrule(lr){8-11}
& \multicolumn{2}{c}{GPT-5.1}
& \multicolumn{2}{c}{\makecell{Qwen3-VL\\8B}}
& \multicolumn{2}{c}{\makecell{Qwen3-VL\\32B}}
& \multicolumn{2}{c}{\makecell{SurgAtlas\\only}}
& \multicolumn{2}{c}{\makecell{SurgAtlas\\+ public}} \\
\cmidrule(lr){2-3} \cmidrule(lr){4-5} \cmidrule(lr){6-7} \cmidrule(lr){8-9} \cmidrule(lr){10-11}
Category & MIS & Open & MIS & Open & MIS & Open & MIS & Open & MIS & Open \\
\midrule
\multicolumn{11}{l}{\textit{Perception \& identification}} \\
\quad Entity existence
& 39.5 & 35.4 & 24.0 & 24.8 & 28.0 & 25.9 & \textbf{49.8} & \textbf{48.0} & 45.9 & 47.2 \\
\quad Entity state
& 30.0 & 35.5 & 19.7 & 24.8 & 25.3 & 28.8 & 33.2 & 37.6 & \textbf{33.6} & \textbf{38.2} \\
\quad Spatial relation
& 28.3 & 28.7 & 18.5 & 17.3 & 22.0 & 22.4 & 36.7 & 33.1 & \textbf{38.9} & \textbf{33.5} \\
\midrule
\multicolumn{11}{l}{\textit{Action \& procedural state}} \\
\quad \makecell[l]{Instrument--tissue\\interaction}
& 43.6 & 42.5 & 23.2 & 21.1 & 32.0 & 30.1 & 40.8 & 42.4 & \textbf{43.8} & \textbf{44.9} \\
\quad Operative action
& 36.6 & 30.6 & 17.5 & 17.4 & 23.0 & 21.6 & \textbf{38.9} & \textbf{34.5} & 38.6 & 32.8 \\
\midrule
\multicolumn{11}{l}{\textit{Operative reasoning}} \\
\quad Maneuver rationale
& \textbf{48.4} & \textbf{53.4} & 22.4 & 18.7 & 30.4 & 31.9 & 40.7 & 38.1 & 40.2 & 39.8 \\
\quad Decision justification
& \textbf{45.5} & \textbf{46.8} & 20.7 & 19.1 & 36.1 & 32.7 & 37.4 & 35.5 & 36.6 & 37.0 \\
\midrule
\multicolumn{11}{l}{\textit{Temporal \& predictive reasoning}} \\
\quad Procedural sequence
& 44.0 & 32.8 & 30.5 & 24.5 & 33.9 & 28.5 & \textbf{48.6} & 40.5 & 48.2 & \textbf{43.5} \\
\quad Next-step prediction
& 37.8 & 33.2 & 21.8 & 24.9 & 24.8 & 27.1 & 34.9 & \textbf{41.6} & \textbf{36.6} & 38.0 \\
\midrule
\multicolumn{11}{l}{\textit{Risk anatomy identification}} \\
\quad \makecell[l]{Risk anatomy\\identification}
& 54.9 & 56.6 & 41.1 & 39.1 & 45.6 & 47.6 & 57.3 & 58.6 & \textbf{57.5} & \textbf{59.7} \\
\midrule
\textbf{Overall}
& 39.6 & 36.4 & 22.5 & 21.7 & 28.5 & 27.2 & \textbf{41.7} & \textbf{39.4} & 41.6 & \textbf{39.4} \\
\bottomrule
\end{tabular}
\end{table*}


\subsection{Fine-grained results.}
The complete fine-grained breakdown of the expert-validated subset $\mathcal{A}^{\text{exp}}$ across all 10 categories is reported in Table~\ref{tab:vqa_finegrained_appendix}, mirroring the broad-category trends in Table~\ref{tab:vqa_expert}. Here we focus on the larger held-out test split of $\mathcal{A}^{\text{qa}}$. Table~\ref{tab:vqa_finegrained_testset} reports per-category LLM-judge accuracy on this split, evaluated on both MIS and open-surgery partitions. The pattern observed on $\mathcal{A}^{\text{exp}}$ holds here at scale: SurgAtlas variants lead $8$ of the $10$ fine-grained categories on both partitions, with the largest absolute margins in perception (entity existence: $+10.3$ MIS / $+12.6$ Open over GPT-5.1) and temporal/predictive reasoning (procedural sequence: $+4.6$ MIS / $+10.7$ Open). GPT-5.1 retains a substantial lead on operative reasoning---maneuver rationale ($48.4$ / $53.4$) and decision justification ($45.5$ / $46.8$)---consistent with the hypothesis that clinical justification benefits disproportionately from frontier-scale pretraining. The two SurgAtlas variants are essentially tied overall ($41.7$ vs.\ $41.6$ MIS; $39.4$ vs.\ $39.4$ Open), but split structurally: SurgAtlas-only is stronger on pure-perception categories while SurgAtlas+public gains on instrument--tissue interaction and temporal/predictive reasoning, where converted public benchmarks provide aligned supervision. Both variants outperform the $32$B Qwen baseline by $\sim 13$ points overall despite using a $4\times$ smaller backbone.

\end{document}